\documentclass[letterpaper, 10 pt, conference]{ieeeconf}    
\IEEEoverridecommandlockouts                                
\usepackage[T1]{fontenc}
\usepackage{graphicx}
\usepackage{amsmath}
\usepackage{comment}
\usepackage{soul}
\usepackage{caption}                                        
\usepackage{multirow}
\let\labelindent\relax                                      
\usepackage{enumitem}                                       
\usepackage[pro]{fontawesome5}                              
\usepackage{subcaption}                                     
\usepackage{float}                                          
\let\mathcalorig\mathcal                                    
\usepackage{calrsfs}
\usepackage{hyperref}
\usepackage[dvipsnames]{xcolor}                             
\usepackage{listings}                                       
\usepackage{amssymb}
\usepackage{amsfonts}                                       
\usepackage{mathrsfs} 

\usepackage{siunitx}                                        

\usepackage{adjustbox}  
\usepackage{array}      

\usepackage{tikz}
\usepackage{tikz-3dplot}
\usetikzlibrary{fit}
\usetikzlibrary{arrows.meta}
\usetikzlibrary{calc,positioning,backgrounds}
\tdplotsetmaincoords{-60}{-35}
\DeclareMathAlphabet\mathbfcal{OMS}{cmsy}{b}{n}

\definecolor{DFKIBlue}{rgb}{0.113, 0.227, 0.560}
\definecolor{DFKIMagenta}{rgb}{0.925, 0.380, 0.623}
\definecolor{DFKIGrey}{rgb}{0.662, 0.647, 0.576}
\definecolor{DFKIGreen}{rgb}{0.415, 0.749, 0.639}
\definecolor{DFKIBlack}{RGB}{6,24,28}
\definecolor{DFKIYellow}{rgb}{1.0, 0.65, 0.0}

\definecolor{YlGnA}{RGB}{255,255,229}
\definecolor{YlGnB}{RGB}{252,254,211}
\definecolor{YlGnC}{RGB}{248,253,193}
\definecolor{YlGnD}{RGB}{239,249,179}
\definecolor{YlGnE}{RGB}{227,244,170}
\definecolor{YlGnF}{RGB}{213,238,161}
\definecolor{YlGnG}{RGB}{195,230,152}
\definecolor{YlGnH}{RGB}{175,222,143}
\definecolor{YlGnI}{RGB}{154,213,135}
\definecolor{YlGnJ}{RGB}{131,203,125}
\definecolor{YlGnK}{RGB}{109,192,115}
\definecolor{YlGnL}{RGB}{85,181,103}
\definecolor{YlGnM}{RGB}{63,169,92}
\definecolor{YlGnN}{RGB}{50,152,80}
\definecolor{YlGnO}{RGB}{38,136,70}
\definecolor{YlGnP}{RGB}{23,123,63}
\definecolor{YlGnQ}{RGB}{9,111,58}
\definecolor{YlGnR}{RGB}{0,98,52}
\definecolor{YlGnS}{RGB}{0,83,47}
\definecolor{YlGnT}{RGB}{0,69,41}

\definecolor{MagmaBlack}{rgb}{0.001462, 0.000466, 0.013866}
\definecolor{MagmaPurple}{rgb}{0.316654, 0.07169, 0.48538}
\definecolor{MagmaMagenta}{rgb}{0.716387, 0.214982, 0.47529}
\definecolor{MagmaOrange}{rgb}{0.9867, 0.535582, 0.38221}
\definecolor{MagmaYellow}{rgb}{0.987053, 0.991438, 0.749504}

\definecolor{secondColor}{RGB}{51, 51, 255}
\definecolor{customDandelion}{RGB}{253, 188, 66}
\definecolor{customGrey}{rgb}{0.7, 0.7, 0.7}
\definecolor{customLightGrey}{rgb}{0.9, 0.9, 0.9}
\definecolor{customBlack}{rgb}{0, 0, 0}

\definecolor{grayclvr}{RGB}{109, 109, 109}
\definecolor{redclvr}{RGB}{216, 44, 44}
\definecolor{blueclvr}{RGB}{53, 94, 255}
\definecolor{greenclvr}{RGB}{36, 131, 25}
\definecolor{brownclvr}{RGB}{161, 93, 31}
\definecolor{purpleclvr}{RGB}{161, 47, 240}
\definecolor{cyanclvr}{RGB}{51, 255, 255}
\definecolor{yellowclvr}{RGB}{255, 255, 64}

\definecolor{codegreen}{rgb}{0,0.6,0}
\definecolor{codegray}{rgb}{0.5,0.5,0.5}
\definecolor{codepurple}{rgb}{0.58,0,0.82}
\definecolor{backcolour}{rgb}{0.95,0.95,0.92}

\definecolor{cubeColor}{RGB}{150, 75, 0}
\definecolor{skyColor}{RGB}{135, 206, 250}
\definecolor{skyLightColor}{rgb}{0.68, 0.85, 0.90}
\definecolor{lightColor}{rgb}{0.968, 0.654, 0.0705}
\definecolor{floorGridColor}{RGB}{143, 227, 136}
\definecolor{floorColor}{RGB}{0, 154, 49}

\colorlet{worldModelLineColor}{DFKIYellow}
\colorlet{worldModelNumberBorderColor}{black}
\colorlet{worldModelNumberColor}{black}

\colorlet{globalMethodColor}{DFKIGreen!85!white}
\colorlet{globalMethodColorLow}{DFKIGreen!10!white}
\colorlet{globalMethodStepColor}{DFKIYellow!80}
\colorlet{globalMethodStepColorLow}{DFKIYellow!10}
\colorlet{globalMethodColor}{skyColor!50!white}
\colorlet{globalMethodColorLow}{skyColor!40!white}
\colorlet{globalMethodStepColor}{floorGridColor}
\colorlet{globalMethodStepColorLow}{floorGridColor}
\colorlet{globalMethodLineColor}{DFKIGrey}
\colorlet{globalMethodBorderColor}{DFKIGrey}
\colorlet{globalMethodStepColorA}{YlGnB}
\colorlet{globalMethodStepColorB}{YlGnD}
\colorlet{globalMethodStepColorC}{YlGnF}
\colorlet{globalMethodStepColorD}{YlGnH}
\colorlet{globalMethodStepColorE}{YlGnJ}

\colorlet{constraintColor}{DFKIYellow}
\colorlet{frameColor}{DFKIGrey}
\colorlet{planeColor}{customLightGrey}
\colorlet{objectColor}{YlGnL}

\colorlet{teapotArrowColor}{DFKIYellow}
\colorlet{parametersColor}{DFKIYellow}
\colorlet{parametersFillColor}{DFKIYellow!10}
\colorlet{parametersTextColor}{black}
\colorlet{observationsColor}{skyColor!80!black}
\colorlet{pipelineColor}{DFKIGrey}
\colorlet{pipelineBoxColor}{DFKIGrey}
\colorlet{sceneBorderColor}{DFKIGrey}

\colorlet{robotSetupLineColor}{DFKIYellow}
\colorlet{robotSetupTextColor}{black}

\newcommand{\ModelName}{Differentiable Neuro-Graphics}

\newcommand{\modelname}{differentiable neuro-graphics}
\newcommand{\ModelAcronym}{DNG}
\newcommand{\MyModel}{\texttt{DNG}}

\newcommand{\MyModelTwo}{\texttt{DNG2}}
\newcommand{\MyModelThree}{\texttt{DNG3}}
\newcommand{\MAP}{MAP}
\newcommand{\GenSixD}{Gen6D}
\newcommand{\OnePosePlusPlus}{OnePose++}

\lstdefinestyle{mystyle}{
    backgroundcolor=\color{backcolour},
    commentstyle=\color{codegreen},
    keywordstyle=\color{magenta},
    numberstyle=\tiny\color{codegray},
    stringstyle=\color{codepurple},
    basicstyle=\ttfamily\footnotesize,
    breakatwhitespace=false,
    breaklines=true,
    captionpos=b,
    keepspaces=true,
    numbers=left,
    numbersep=5pt,
    showspaces=false,
    showstringspaces=false,
    showtabs=false,
    tabsize=2
}

\newcolumntype{R}[2]{%
    >{\adjustbox{angle=#1,lap=\width-(#2)}\bgroup}%
    l%
    <{\egroup}%
}

\newcommand*\rot{\multicolumn{1}{R{-20}{1em}}}

\newcommand*\rotb{\multicolumn{1}{R{-30}{1em}}}

\DeclareMathAlphabet{\pazocal}{OMS}{zplm}{m}{n}

\DeclareMathOperator*{\argmin}{arg\,min}
\DeclareMathOperator*{\argmax}{arg\,max}

\newcommand{\ObjectArg}{k}
\newcommand{\NumObjects}{K}
\newcommand{\EstimatedNumObjects}{\widehat{K}}
\newcommand{\Parameter}{\theta}
\newcommand{\ScalarScale}{s}
\newcommand{\Parameters}{{\boldsymbol{\Parameter}}}

\newcommand{\ScalarPosition}{p}

\newcommand{\Position}{\mathbf{\ScalarPosition}}
\newcommand{\RGBD}{\textrm{RGBD}}
\newcommand{\RGB}{\textrm{RGB}}
\newcommand{\Image}{\mathbf{I}}
\newcommand{\DepthVariable}{D}          
\newcommand{\depthVariable}{d}          
\newcommand{\Depth}{\mathbf{\DepthVariable}}
\newcommand{\Mask}{\mathbf{M}}
\newcommand{\LBFGS}{\mbox{L-BFGS}}
\newcommand{\Vertices}{\mathbf{V}}

\newcommand{\Pose}{\textbf{T}}
\newcommand{\BoundingBox}{\mathbf{B}}
\newcommand{\EstimatedBoundingBoxes}{\widehat{\mathbf{B}}}
\newcommand{\Mesh}{\mathbf{O}}
\newcommand{\SAM}{\textrm{SAM}}
\newcommand{\Intensity}{i}
\newcommand{\Material}{\mathbf{m}}
\newcommand{\Light}{\boldsymbol{\ell}}
\newcommand{\JAX}{\texttt{JAX}}
\newcommand{\Model}{\pazocal{G}}
\newcommand{\Reals}{\mathbb{R}}
\newcommand{\NumVertices}{N}
\newcommand{\NumCageVertices}{\mathrm{N_c}}

\newcommand{\RotationGroup}{\mathrm{SO}(3)}
\newcommand{\CLVRPOSE}{\texttt{CLEVR-POSE}}
\newcommand{\FEWSOL}{\texttt{FewSOL}}
\newcommand{\MOPED}{\texttt{MOPED}}
\newcommand{\LINEMOD}{\texttt{LINEMOD-OCCLUDED}}
\newcommand{\EFFICIENTDET}{\texttt{EFFICIENTDET-D7}}
\newcommand{\YCB}{\texttt{YCB}}
\newcommand{\SSD}{\texttt{SSD512}}
\newcommand{\Pointcloud}{\mathbf{C}}
\newcommand{\PointcloudCoordinate}{c}
\newcommand{\CameraIntrinsics}{\mathbf{K}}
\newcommand{\FocalLength}{f}
\newcommand{\Center}{c}
\newcommand{\Scale}{\mathbf{s}}
\newcommand{\PointcloudArg}{i}
\newcommand{\Probability}[1]{p\left(#1\right)}
\newcommand{\Cardinality}[1]{|#1|}
\newcommand{\MeanAbsoluteErrorVariable}{\pazocal{E}}
\newcommand{\Laplace}{\mathrm{Lap}}
\newcommand{\TruncatedNormal}{\mathcalorig{TN}}
\newcommand{\LogNormal}{\mathrm{LogN}}
\newcommand{\Ellipsoid}{e}
\newcommand{\Ambient}{{\alpha_k}}
\newcommand{\Diffuse}{{\beta_k}}
\newcommand{\Specular}{{\gamma_k}}
\newcommand{\Shininess}{{\delta_k}}
\newcommand{\Color}{{\boldsymbol{\epsilon}_k}}
\newcommand{\ImageWidth}{U}
\newcommand{\ImageHeight}{V}
\newcommand{\ImageWidthArg}{u}
\newcommand{\ImageHeightArg}{v}
\newcommand{\Stdv}{\sigma}
\newcommand{\Mean}{\mu}
\newcommand{\LineDistance}{t}
\newcommand{\Loss}{\pazocal{L}}
\newcommand{\Curvature}{c}
\newcommand{\Barrier}{b}
\newcommand{\DifferentiableRenderer}{\pazocal{R}}
\newcommand{\CameraOrigin}{\mathbf{o}}
\newcommand{\CameraExtrinsics}{[\, \mathbf{R}_{\ObjectArg} |\, \mathbf{t}_{\ObjectArg}]}

\newcommand{\Floor}{f}
\newcommand{\SceneVariable}{s}

\newcommand{\Direction}{\mathbf{d}}
\newcommand{\Volume}{v}
\newcommand{\Smooth}{t}
\newcommand{\Angle}{\theta}
\newcommand{\Pattern}{\mathbf{p}}

\newcommand{\BoundingBoxes}{\BoundingBox_{\ObjectArg}}
\newcommand{\Masks}{\Mask_{\ObjectArg}}
\newcommand{\Poses}{\Pose_{\ObjectArg}}
\newcommand{\Meshes}{\Mesh_{\ObjectArg}}
\newcommand{\Materials}{\Material_{\ObjectArg}}

\newcommand{\EstimatedMaterials}{\widehat{\Material}_{\ObjectArg}}
\newcommand{\EstimatedMasks}{\widehat{\Mask}_{\ObjectArg}}
\newcommand{\EstimatedPoses}{\widehat{\Pose}_{\ObjectArg}}
\newcommand{\EstimatedMeshes}{\widehat{\Mesh}_{\ObjectArg}}

\newcommand{\LightIntensity}{\Light^{\Intensity}}
\newcommand{\LightPosition}{\Light^{\ScalarPosition}}
\newcommand{\SceneParameters}{\boldsymbol{\theta}^{\text{s}}_{\ObjectArg}}
\newcommand{\NeuralParameters}{\boldsymbol{\theta}^{\text{n}}}
\newcommand{\Positions}{\Position_{\ObjectArg}}
\newcommand{\Scales}{\Scale_{\ObjectArg}}
\newcommand{\FocalLengthX}{\FocalLength_x}
\newcommand{\FocalLengthY}{\FocalLength_y}
\newcommand{\ImageCenter}{\Center}
\newcommand{\ImageCenterX}{\ImageCenter_x}
\newcommand{\ImageCenterY}{\ImageCenter_y}
\newcommand{\PointcloudX}{c_{i}^x}
\newcommand{\PointcloudY}{c_{i}^y}
\newcommand{\PointcloudZ}{c_{i}^z}
\newcommand{\PositionsX}{\ScalarPosition_{\ObjectArg}^x}
\newcommand{\PositionsY}{\ScalarPosition_{\ObjectArg}^y}
\newcommand{\PositionsZ}{\ScalarPosition_{\ObjectArg}^z}
\newcommand{\ScalesX}{\ScalarScale_{\ObjectArg}^x}
\newcommand{\ScalesY}{\ScalarScale_{\ObjectArg}^y}
\newcommand{\ScalesZ}{\ScalarScale_{\ObjectArg}^z}
\newcommand{\Pointclouds}{\boldsymbol{\PointcloudCoordinate}_{\ObjectArg}}
\newcommand{\EllipsoidPoints}{\Ellipsoid_{\PointcloudArg}}
\newcommand{\EllipsoidsPoints}{\Ellipsoid_{\PointcloudArg, \ObjectArg}}

\newcommand{\EllipsoidParameters}{\boldsymbol{\Parameter}^{\textrm{\Ellipsoid}}_\ObjectArg}
\newcommand{\EstimatedEllipsoidParameters}{\hat{\boldsymbol{\Parameter}}^{\textrm{\Ellipsoid}}_\ObjectArg}

\newcommand{\PointcloudMeanPosition}{\boldsymbol{\Mean}^{\ScalarPosition}_\ObjectArg}
\newcommand{\PointcloudMeanPositionX}{\Mean^{\ScalarPosition,x}_\ObjectArg}
\newcommand{\PointcloudMeanPositionY}{\Mean^{\ScalarPosition,y}_\ObjectArg}
\newcommand{\PointcloudMeanPositionZ}{\Mean^{\ScalarPosition,z}_\ObjectArg}
\newcommand{\PointcloudStdvPosition}{\boldsymbol{\Stdv}^{\ScalarPosition}_\ObjectArg}
\newcommand{\PointcloudStdvPositionX}{\Stdv^{\ScalarPosition,x}_\ObjectArg}
\newcommand{\PointcloudStdvPositionY}{\Stdv^{\ScalarPosition,y}_\ObjectArg}
\newcommand{\PointcloudStdvPositionZ}{\Stdv^{\ScalarPosition,z}_\ObjectArg}
\newcommand{\PositionPriorScale}{\boldsymbol{\sigma}^{p}}
\newcommand{\SizePriorScale}{\sigma^{s}}

\newcommand{\PointcloudsPoints}{{\PointcloudCoordinate}_{\PointcloudArg, \ObjectArg}}
\newcommand{\ObservedPointcloudsPointsX}{{\hat{\PointcloudCoordinate}}_{\PointcloudArg, \ObjectArg}^x}
\newcommand{\ObservedPointcloudsPointsY}{{\hat{\PointcloudCoordinate}}_{\PointcloudArg, \ObjectArg}^y}
\newcommand{\ObservedPointcloudsPointsZ}{{\hat{\PointcloudCoordinate}}_{\PointcloudArg, \ObjectArg}^z}
\newcommand{\ObservedPointcloudsPoints}{\hat{\PointcloudCoordinate}_{\PointcloudArg, \ObjectArg}}
\newcommand{\MinDepth}{\depthVariable_{\textrm{min}}}
\newcommand{\MaxDepth}{\depthVariable_{\textrm{max}}}
\newcommand{\RenderedImage}{\widehat{\Image}}
\newcommand{\RenderedDepth}{\widehat{\Depth}}
\newcommand{\RenderedMask}{{\widehat{\mathbf{M}}^{\text{R}}}}
\newcommand{\RenderedMasks}{{\RenderedMask}_{\ObjectArg}}

\newcommand{\SceneImageWeight}{w^{i}}
\newcommand{\SceneDepthWeight}{w^{d}}
\newcommand{\SceneMasksWeight}{w^{m}}

\newcommand{\ImageErrors}{\MeanAbsoluteErrorVariable^{\text{i}}_{\ObjectArg}}
\newcommand{\DepthErrors}{\MeanAbsoluteErrorVariable^{\text{d}}_{\ObjectArg}}
\newcommand{\MasksErrors}{\MeanAbsoluteErrorVariable^{\text{m}}_{\ObjectArg}}

\newcommand{\MaskedImage}{\Image^{m}}
\newcommand{\MaskedDepth}{\Depth^{m}}
\newcommand{\MaskedRenderedImage}{\RenderedImage^{m}}
\newcommand{\MaskedRenderedDepth}{\RenderedDepth^{m}}
\newcommand{\MaskedImages}{\MaskedImage_{\ObjectArg}}
\newcommand{\MaskedDepths}{\MaskedDepth_{\ObjectArg}}
\newcommand{\MaskedRenderedDepths}{\MaskedRenderedDepth_{\ObjectArg}}
\newcommand{\MaskedRenderedImages}{\MaskedRenderedImage_{\ObjectArg}}
\newcommand{\SceneLoss}{\Loss^{\SceneVariable}}

\newcommand{\FloorMaterial}{\Material^{\Floor}}
\newcommand{\OptimizedSceneParameters}{\boldsymbol{\hat{\theta}}^{\text{s}}_{\ObjectArg}}

\newcommand{\BarrierLoss}{\Loss^{\Barrier}}

\newcommand{\VolumeLoss}{\Loss^{\Volume}}
\newcommand{\VolumeLosses}{\VolumeLoss_{\ObjectArg}}
\newcommand{\VolumeLossWeight}{w^{\Volume}}
\newcommand{\SceneVolumes}{\Volume_{\ObjectArg}^{s}}
\newcommand{\MeshVolumes}{\Volume_{\ObjectArg}^{m}}

\newcommand{\SmoothDepthLoss}{\Loss^{\Smooth}}
\newcommand{\SmoothDepthLosses}{\SmoothDepthLoss_{\ObjectArg}}
\newcommand{\SmoothDepthLossWeight}{w^{\Smooth}}

\newcommand{\SmoothMeshLoss}{\Loss^{m}}
\newcommand{\SmoothMeshLosses}{\SmoothMeshLoss_{\ObjectArg}}
\newcommand{\SmoothMeshLossWeight}{w^{m}}

\newcommand{\MeshLoss}{\Loss^{o}}

\newcommand{\ShapeParameters}{\boldsymbol{\Parameter}^{\text{o}}_{\ObjectArg}}
\newcommand{\OptimizedShapeParameters}{\hat{\boldsymbol{\Parameter}}^{\text{o}}_{\ObjectArg}}

\newif\ifhighlight

\highlightfalse

\ifhighlight
    \newcommand{\highlightOne}[1]{{%
        \colorlet{foo}{yellow!40}%
        \sethlcolor{foo}\hl{#1}}%
    }
\else
  \newcommand{\highlightOne}[1]{#1}
\fi

\title{\LARGE \bf
  Differentiable Inverse Graphics \\ for Zero-shot Scene Reconstruction and Robot Grasping
}

\author{Octavio Arriaga$^{1}$, Proneet Sharma$^{1}$, Jichen Guo, Marc Otto$^{1}$, Siddhant Kadwe, Rebecca Adam$^{1}$ \\
$^{1}$Robotics Innovation Center, DFKI GmbH
\thanks{This work has been submitted to the IEEE for possible publication. Copyright may be transferred without notice, after which this version may no longer be accessible.}}

\begin{document}
    \maketitle
    \thispagestyle{empty}
    \pagestyle{empty}

    \begin{abstract}
        Operating effectively in novel real-world environments requires robotic systems to estimate and interact with previously unseen objects.
Current state-of-the-art models address this challenge by using large amounts of training data and test-time samples to build black-box scene representations.
In this work, we introduce a \modelname~model that combines neural foundation models with physics-based differentiable rendering to perform zero-shot scene reconstruction and robot grasping without relying on any additional 3D data or test-time samples.
Our model solves a series of constrained optimization problems to estimate physically consistent scene parameters, such as meshes, lighting conditions, material properties, and 6D poses of previously unseen objects from a single RGBD image and bounding boxes.
We evaluated our approach on standard model-free few-shot benchmarks and demonstrated that it outperforms existing algorithms for model-free few-shot pose estimation.
Furthermore, we validated the accuracy of our scene reconstructions by applying our algorithm to a zero-shot grasping task.
By enabling zero-shot, physically-consistent scene reconstruction and grasping without reliance on extensive datasets or test-time sampling, our approach offers a pathway towards more data efficient, interpretable and generalizable robot autonomy in novel environments.

    \end{abstract}

    \section{Introduction}\label{sec:introduction}
        \begin{figure}[t!]
    \centering
    \scalebox{0.58}{%
        \input{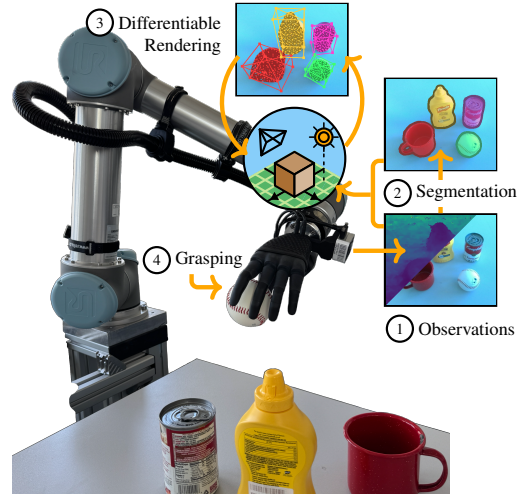}
    }
    \caption{(1) The system first observes the scene with an \RGBD~camera.
        (2) The \RGB~image is segmented using a foundation model and an object detector to obtain object masks, which are then combined with the observations to initialize scene geometry.
        (3) A physics-based differentiable renderer iteratively refines the system's internal world model, encompassing meshes, lights, and materials, by comparing rendered images to real observations.
    (4) The optimized 3D scene enables accurate grasping of novel objects.~\label{fig:WorldModel}}
    \vspace{-0.75cm}
\end{figure}
\begin{figure*}[t!]
    \centering
    \vspace{0.1cm}
    \scalebox{0.66}{
    \input{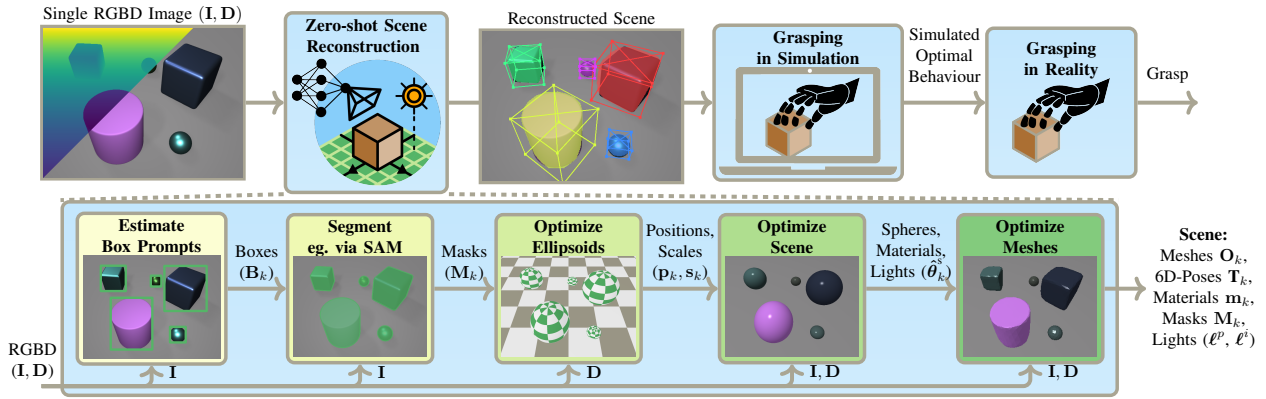}
}
    \caption{\ModelName~for zero-shot scene reconstruction and robotic manipulation.
        Starting from an $\RGBD$ image and bounding box prompts, the system uses the segmentation model SAM to initialize the object masks.
    It then initializes a 3D scene by performing a robust probabilistic estimation of object shapes using ellipsoidal primitives.
    Subsequently, a mesh optimization stage refines the mesh vertices through a cage-based deformation model.
    The resulting scene representation includes meshes, poses, materials, masks, and lighting conditions.
    Finally, the reconstructed scene is used in simulation to find an optimal grasp, which is then performed in reality using the robotic system.~\label{fig:GlobalMethodPipeline}}
    \vspace{-0.5cm}
\end{figure*}
A striking feature of human visual cognition is its ability to predict the state of the world with limited data~\cite{lake2015human}.
Humans are capable of determining the shape, material, and position of completely novel objects by just observing them once~\cite{goldstein2015cognitive}.
Enabling robotic systems to solve these tasks could provide them with the necessary autonomy to operate in unstructured novel environments, such as in search and rescue or domestic settings.
Current vision models address these issues by optimizing large-scale neural networks with vast amount of data, that can be adapted to unseen environments and tasks.
This paradigm has provided enormous contributions to the field of computer vision, leading to foundation models in instance segmentation~\cite{kirillov2023segment}, feature extraction~\cite{oquab2023dinov2}, and vision-language-action models~\cite{black2024pi0}.
Despite these contributions, directly applying deep learning models to robot perception-action tasks still presents several open challenges.
One major hurdle is \textit{sample complexity}, as foundation vision models often require millions or even billions of labeled training samples, making the acquisition of similar datasets for 3D robot perception and robot grasping exceptionally expensive and time-consuming.
Another significant challenge is the lack of \textit{explainability} inherent in many deep learning models, thereby hindering trust and safety when deploying them in critical robotic applications.
While hallucinations in large language models can often be remedied by search and fact checking at a later time, hallucinations in high-torque systems carry the immediate risk of long-term injury or death.
\highlightOne{Furthermore, contrary to the premise of few-shot learning, many state-of-the-art deep learning models for few-shot pose estimation still depend on \textit{complex test-time data}, requiring either hundreds of images~\mbox{\cite{liu2022gen6d, he2022fs6d}} or several 6D-annotated frames~{\cite{wen2023foundationpose}}.
    This reliance on costly test-time data, combined with large background training, deviates from the challenge of learning from minimal information.
    This creates a significant practical bottleneck, limiting robotic applications which demand immediate interaction with novel objects.}
\highlightOne{On the other hand, the emergent field of physics-based differentiable rendering offers greater interpretability but presents its own set of challenges.
    Specifically, these methods typically require idealized conditions, such as dozens of views, camera poses, or accurate mask information, which are also often unavailable in real robotic scenarios.

    Both deep learning and differentiable graphics, therefore, impose data requirements misaligned with the goal of enabling autonomous robots to operate from minimal, real-world observations.
    Early work in few-shot learning~{\cite{lake2019omniglot}} highlighted that the key challenge is not just learning from vast datasets, but learning from minimal data by incorporating physical and cognitive knowledge.
    In line with this cognitive perspective, our work introduces a physically grounded, sequential inverse optimization framework designed to learn from such minimal observations. 
    Our core contribution, illustrated in Figure~{\ref{fig:WorldModel}}, enables robots to reconstruct and grasp completely unseen objects from a single $\RGBD$ image, without requiring multi-view captures, 6D pose labels, and expensive collection of 3D training data.
    The global pipeline of this framework is shown in Figure~{\ref{fig:GlobalMethodPipeline}}, and its novel contributions are:}
\begin{itemize}[leftmargin=3mm, noitemsep]
    \item \highlightOne{\textbf{Zero-Shot Reconstruction Pipeline:} We propose a novel multi-stage optimization pipeline that replaces traditional training and reconstructs unseen objects from a single RGBD image without any 3D training data. This coarse-to-fine strategy  surpasses model-free few-shot pose estimation models on standard 6D pose benchmarks and is validated on a zero-shot robotic grasping task.}

    \item \highlightOne{\textbf{Robust Optimization Methodology:} This work introduces a new probabilistic ellipsoid estimation that uses physically-consistent priors to account for sensor noise. This serves as a critical, robust initialization for our efficient Quasi-Newton optimization, allowing the pipeline to avoid local minima, which we validate in our results.}

    \item \highlightOne{\textbf{Differentiable Graphics Engine:} We developed a novel differentiable ray tracer for triangular meshes using only \JAX-primitives.
    This enables rapid research prototyping, high portability, and integrates seamlessly with optimization and deep learning libraries.
    Furthermore, it features a new differentiable soft-mask function to address the zero-gradient problem inherent in rendering binary masks.}
\end{itemize}
\highlightOne{These contributions hold the potential to advance data-efficient and physically grounded robot perception and action, enabling their application in unstructured environments.}

    \section{Related work}\label{sec:related_work}
        This paper focuses on methodological contributions to model-free few-shot pose estimation; accordingly, this literature review prioritizes relevant perception techniques.
While robotic grasping validates and extends our approach, the extensive literature specific to grasping itself is excluded from this section's scope because it represents a downstream application beyond our core perception focus.
Few-shot pose estimation methods address the large sample complexity as well as the limited generalization to new objects often encountered in category-specific pose estimation models~\cite{he2022fs6d, liu2022gen6d, he2022onepose++, goodwin2022, lin2024sam}.
While these algorithms have demonstrated impressive performance, their generalization depends on extensive pre-training, as well as additional images for each new object during test-time~\cite{liu2022gen6d, he2022onepose++, wen2023foundationpose}.
Specifically, methods like FS6D required 800K training samples, Gen6D~\cite{liu2022gen6d} approximately 2.6M samples, and LatentFusion 960M~\cite{park2020latentfusion}.
\highlightOne{Furthermore, most few-shot pose estimation models~\mbox{\cite{liu2022gen6d, he2022onepose++, wen2023foundationpose}} require an additional data-collection step to build 3D object representations.
    While FoundationPose uses fewer reference images (typically 4-16), those require ground-truth 6D pose annotations and an initial mask.
    OnePose++ and Gen6D both rely on COLMAP~{\cite{schonberger2016structure}} for 3D reconstruction, which requires capturing multi-view videos with sufficient texture to extract reliable features and build object models.}
In contrast, recent work in physically-based differentiable rendering has developed regularization terms and loss functions for accurate 3D object reconstruction from multi-view images and object masks without explicitly requiring a feature extraction model~\cite{nicolet2021large, ravi2020pytorch3d, liu2019soft}.
Moreover, recent advances in neural rendering techniques have led to methods like NeRF-Pose~\cite{li2023nerf}, which propose weakly-supervised training pipelines for estimating 6D poses.
Prior differentiable rendering methods often optimize 3D structures without pre-defined models but typically neglect few-shot learning or robotic system challenges.

    \section{Hybrid Neuro-Graphics Model}\label{sec:methodology}
        Physics-based rendering simulates the causal physical process of image formation, where the final appearance is dictated by object properties like geometry, material, and lighting conditions.
We propose the \ModelName~(\ModelAcronym) model to solve the inverse problem of inferring these scene properties from a single image, via a cascade of constrained optimization problems.
Let $\NumObjects$ denote the varying number of objects per image, and let $U,V$ denote the image pixel width and height, respectively.
Then our {\ModelAcronym} model $\Model$ takes as input a color image $\Image \in \Reals^{U \times V \times 3}$, and a depth image $\Depth \in \Reals^{U \times V}$.
As output for each object $k \in \{1, \hdots , K\}$, {\ModelAcronym} reconstructs the mesh $\Meshes \in \Reals^{N \times 3}$ specified by $\NumVertices$ vertex locations, the pose $\Poses = \CameraExtrinsics$ with rotation matrix $\mathbf{R}_k \in \RotationGroup$ and translation vector $\mathbf{t}_k \in \mathbb{R}^{3 \times 1}$, the mask $\Masks \in \Reals^{U \times V}$, the lighting conditions $\Light = [ \LightPosition, \LightIntensity]$ consisting of the light's position $\LightPosition$ and intensity $\LightIntensity$, and the material $\Materials = [\Ambient \Diffuse, \Specular, \Shininess, \Color]$ with the properties ambient $\Ambient$, diffuse $\Diffuse$, specular $\Specular$, and shininess $\Shininess$, and color $\Color \in \Reals^{3}$
\begin{equation}
    \EstimatedMeshes, \EstimatedPoses, \EstimatedMasks, \EstimatedMaterials, \widehat{\Light}= \Model(\Image, \Depth).
\end{equation}
Our model $\Model$ performs its predictions in four steps: object detection and segmentation, robust ellipsoid estimation, differentiable scene rendering, and differentiable mesh rendering, as shown in Fig.~\ref{fig:GlobalMethodPipeline} and detailed in the next subsections.

        \subsection{Object Detection and Segmentation}
            From an input image, our model first uses an object detector to estimate $\EstimatedNumObjects$ bounding boxes $\BoundingBoxes$.
\begin{align}
    \EstimatedBoundingBoxes = \mathrm{OD}(\Image).
\end{align}
We then input these bounding boxes into the foundation model $\SAM$.
The model utilizes its original pre-trained weights $\NeuralParameters$ to predict the segmentation masks $\Masks$
\begin{equation}
    \EstimatedMasks = \SAM(\Image, \EstimatedBoundingBoxes, \NeuralParameters).
\end{equation}
For our evaluations we used the provided bounding boxes within each dataset; however, for our robotic experiments we used two object detectors.
Specifically the \EFFICIENTDET~model~\cite{tan2020efficientdet} pre-trained in COCO and a custom \SSD~model~\cite{liu2016ssd} fine-tuned on VOC objects.

        \subsection{Robust Ellipsoid Estimation}
            The second step uses the predicted masks $\EstimatedMasks$ and the depth image to compute an initial estimate of the 3D position $\Positions =[\PositionsX, \PositionsY, \PositionsZ]$ and size $\Scales =[\ScalesX, \ScalesY, \ScalesZ]$ of each object $\ObjectArg$.
We begin by projecting $\Depth \in \Reals^{\ImageWidth \times \ImageHeight}$ into a pointcloud $\Pointcloud\in\Reals^{\ImageWidth \cdot \ImageHeight \times 3}$ using the camera intrinsics $\CameraIntrinsics \in \Reals^{3 \times 3}$ with focal lengths $(\FocalLengthX, \FocalLengthY)$ and image center $(\ImageCenterX, \ImageCenterY)$ such that for all $\ImageWidthArg \in \{ 1, \ldots ,\ImageWidth \}$, and $\ImageHeightArg \in \{ 1, \ldots ,\ImageHeight \}$
\begin{align}
    \begin{bmatrix}\PointcloudX \\ \PointcloudY\\ \PointcloudZ\end{bmatrix} =
        \Depth(u,v) \cdot \CameraIntrinsics^{-1}
    \begin{bmatrix}
        u \\
        v \\
        1
    \end{bmatrix}, \text{ }
    \CameraIntrinsics=
        \begin{bmatrix}
            \FocalLengthX & 0 & \ImageCenterX\\
            0 & \FocalLengthY & \ImageCenterY\\
            0 & 0 & 1\\
        \end{bmatrix},
\end{align}
where the map from each pixel $(\ImageWidthArg, \ImageHeightArg)$ to the pointcloud index $\PointcloudArg$ is given by $\PointcloudArg=(\ImageHeightArg - 1) \cdot \ImageWidth + \ImageWidthArg$.
We then partition the pointcloud  $\Pointcloud$ into \highlightOne{multiple} parts $\Pointclouds$ by extracting the points that lie within each of the masks $\EstimatedMasks$.
For each of these $\EstimatedNumObjects$ individual pointclouds we jointly model the position $\Positions$ and the size $\Scales$ to define an ellipsoid surface $\EllipsoidPoints$ for each point $\PointcloudArg$
\begin{align}
    \EllipsoidsPoints =
        \frac{(\ObservedPointcloudsPointsX - \PositionsX)^2}{\ScalesX} +
        \frac{(\ObservedPointcloudsPointsY - \PositionsY)^2}{\ScalesY} +
        \frac{(\ObservedPointcloudsPointsZ - \PositionsZ)^2}{\ScalesZ}.
        \label{eq:ellipsoid}
\end{align}
We estimate these parameters using a maximum a posteriori (\MAP) approach.
To account for the inherent noise in depth image measurements, and residual background points included due to the over-segmentation of object masks, we employ robust prior probability distributions.
Specifically, we model the prior probability distribution of the object's location using the Laplace distribution ($\Laplace$) and its depth size using the Log-normal distribution ($\LogNormal$), both of which are long-tailed distributions designed to handle outliers and larger noise variations.
Furthermore, the object's width and height use a truncated normal distribution ($\TruncatedNormal$) within the positive range $[\MinDepth, \MaxDepth]$.
Both the $\LogNormal$ and $\TruncatedNormal$ priors ensure that the object sizes remain physically plausible by enforcing strictly positive values.
As parameters to our distributions we define the position $\PositionPriorScale$ scale and size $\SizePriorScale$ scale to be \qty{0.1}{\metre} for all values.
Moreover, we use the mean pointcloud position $\PointcloudMeanPosition = [\PointcloudMeanPositionX, \PointcloudMeanPositionY, \PointcloudMeanPositionZ]$ and standard deviation $\PointcloudStdvPosition = [\PointcloudStdvPositionX, \PointcloudStdvPositionY, \PointcloudStdvPositionZ]$ to respectively define the first moment for the position and size distributions.
Thus the priors read: 
\begin{align}
    \Probability{\ScalesX} &= \TruncatedNormal(2 \PointcloudStdvPositionX, \SizePriorScale, \MinDepth, \MaxDepth), \\
    \Probability{\ScalesY} &= \TruncatedNormal(2 \PointcloudStdvPositionY, \SizePriorScale, \MinDepth, \MaxDepth), \\
    \Probability{\ScalesZ} &= \LogNormal(\PointcloudStdvPositionZ, \SizePriorScale), \\
    \Probability{\Positions^{\phantom{a}}} &= \Laplace(\PointcloudMeanPosition, \PositionPriorScale).
\end{align}
Then our likelihood function estimates the probability of the observed point cloud $\ObservedPointcloudsPoints$ given the hypothesis parametrized by $\EllipsoidParameters = [\Positions, \Scales]$ satisfying the ellipsoid surface equation:
\begin{equation}
    p(\ObservedPointcloudsPoints \vert \EllipsoidParameters) = \Laplace \left( 1 \Big| \EllipsoidsPoints, \EllipsoidParameters \right).
\end{equation}
Finally we maximize the posterior density using \LBFGS:
\begin{equation}\label{eq:ellipsoidMathematicalOptimization}
    \EstimatedEllipsoidParameters = \argmax_{\EllipsoidParameters}
        \left\{
            \prod_{\PointcloudArg=1}^{\Cardinality{\Pointclouds}}
                \Probability{\PointcloudsPoints \vert \EllipsoidParameters}
            \hspace{-0.35cm}
                \prod_{j \in \{x, y, z\}}
            \hspace{-0.35cm}
            \Probability{\ScalarScale^{j}_{\ObjectArg}}
            \Probability{\ScalarPosition^{j}_{\ObjectArg}}
        \right\}.
\end{equation}

            \subsection{Differentiable Scene Rendering}\label{sec:scene_rendering}
            The third step uses our differentiable render $\DifferentiableRenderer$ to optimize the scene parameters $\SceneParameters$ using spherical shapes.
These parameters include the lighting, object position, scales, and materials $\SceneParameters = [\LightPosition, \LightIntensity, \Positions, \Scales, \Materials, \FloorMaterial]$.
We initialized the positions $\Positions$ and scales $\Scales$ parameters with the values in $\EstimatedEllipsoidParameters$, estimated in the previous step.
For initialization, we drew the light position and intensity random values from a uniform distribution and we obtained initial material colors by computing the mean of the masked RGB images $\EstimatedMasks$, and the remaining material properties $\Ambient, \Diffuse, \Specular$ were all set to the value of 0.1, and $\Shininess $ to 100.

\begin{figure}[b]
    \centering
    \scalebox{0.8}{
    \begin{tikzpicture}
    [
        tdplot_main_coords,
        ultra thick,
        font=\large,
        >=Stealth,
        my dashed/.style={dashed, thick, ->, shorten >=-15pt, shorten <=-15pt, every node/.append style={font=\footnotesize}},
        myBox/.style={thin, planeColor},
        myObject/.style={objectColor, every node/.append style={fill, circle, inner sep=0pt, minimum size=#1*3.5pt, anchor=center, outer sep=0pt}},
        myVector/.style={ultra thick, constraintColor, line cap=round},
    ]
    \draw [frameColor] (0,4,0) -- (0,7,0) (-2,7,0) -- (2,7,0);
    \coordinate (o) at (0,0,0);
    \coordinate (oshift) at (-1,-2.5,-3.51);
    \path [draw=frameColor, fill=planeColor, opacity=0.8, text opacity=1] (-1.5,4,1.75) coordinate (a) -- ++(0,0,-3.5) coordinate (b) -- ++(3,0,0) coordinate (c) -- ++(0,0,3.5) coordinate (d) -- cycle node [pos=.95, above, sloped, anchor=south west] {
 } ;
    \draw [frameColor] (-2,0,0) -- (2,0,0) (0,0,0) -- (0,4,0) (0,0,0) -- (0,0,2);
    \draw [ultra thick, ->, every node/.style={font=\normalsize, inner sep=0pt}] (o) node [anchor=north west] {$c$} (o) edge node [pos=1, anchor=north east] {$z_w$} ++(0,1,0) edge node [pos=1, anchor=north] {$y_c$} ++(0,0,1) -- ++(1,0,0) node [anchor=north west] {$x_c$};
    \draw [ultra thick, ->, every node/.style={font=\normalsize, inner sep=0pt}] (oshift) node [yshift=0.2cm, anchor=north east] {$w$} (oshift) edge node [pos=1, anchor=north east] {$x_w$} ++(0,-1,0) edge node [pos=1, xshift=0.1cm, anchor=north west] {$z_w$} ++(0,0,-1) -- ++(1,0,0) node [anchor=north west] {$y_w$};
    \draw [myBox] (a) ++(0,0,-2.25) coordinate (p1) -- ++(2.25,0,0) coordinate (p2) -- ++(0,0,-1.25) coordinate (p3);
    \path (p2) ++(-.125,0,0) coordinate (q2) ++(0,0,-.125) coordinate (r2);
    \draw [myObject=3] ($($(a)!1/2!(b)$)+($(q2)-(p1)$)$) coordinate (s2) (r2) node (d1) {};
    \scoped[on background layer]{\draw [myObject=3.5, opacity=0.0] ($($1.75*($(s2)-(0.10,4,0)$)$)+(0,7,0)$) -- ++($1.75*($(r2)-(s2)$)$) node (d2) {};}
    \draw [->,myVector] (o) -- node[align=center, above, xshift=0.15cm, yshift=0.15cm]{$\LineDistance \Direction$}(d1.center);
    \scoped[on background layer]{\draw [myVector, dashed] (d1.center) -- (d2.center);}
    \shade [ball color=objectColor] (d2.center) node[align=center, above, yshift=0.25cm]{$\Positions=\LineDistance \Direction + \CameraOrigin$}circle [radius=0.35cm];
    \draw[<-, ultra thick, constraintColor](o) -- node[align=center, above, xshift=-0.15cm]{$\CameraOrigin$}(oshift);
\end{tikzpicture}
}
    \caption{\protect\highlightOne{Line equality constraint, constraining the $\Reals^{3}$ search space to a single dimension $\LineDistance$ along $d$.}}\label{fig:lineEquality}
\end{figure}
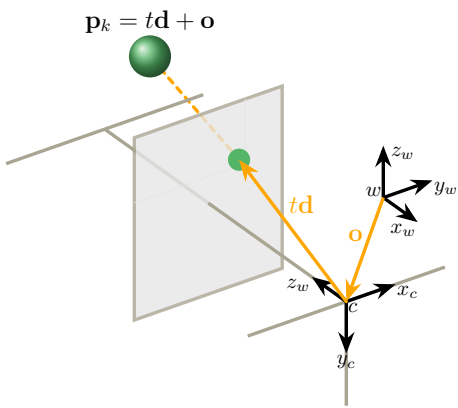

\highlightOne{Throughout the optimization procedure, we enforce physical plausibility by constraining materials to Phong reflection model boundaries~{\cite{bui1975}}. We also introduce an optional 1D line constraint for object positions $\Positions$ (Figure~{\ref{fig:lineEquality}}), which reduces parameters per object from 3 to 1 by optimizing only the distance $\LineDistance$. This step serves as a refinement, offering a minor improvement in convergence stability (Sec.~\mbox{\ref{sec:results_clvr_pose}}).}
We enforce the inequality constraints using the barrier function $\BarrierLoss$ with curvature $\Curvature$ and minimum ($x_{\min}$) and maximum ($x_{\max}$) bounds:
\begin{align}
    \BarrierLoss = \exp(-\Curvature (x - x_{\min})) + \exp(\Curvature (x - x_{\max})).
\end{align}
Having defined all scene variables and constraints, we proceed to define the scene loss function $\SceneLoss$.
Our differentiable renderer $\DifferentiableRenderer$ takes as input the scene parameters $\SceneParameters$, and outputs estimates for the image, $\RenderedImage$, depth $\RenderedDepth$ and masks $\RenderedMasks$:
\begin{align}
    \RenderedImage, \RenderedDepth, \RenderedMasks = \DifferentiableRenderer(\SceneParameters).
\end{align}
For each object $\ObjectArg$, we use the neural masks $\EstimatedMasks$ to mask both the image and depth inputs ($\MaskedImages$, $\MaskedDepths$), and the rendered outputs ($\MaskedRenderedImages$, $\MaskedRenderedDepths$).
Thus our scene loss $\SceneLoss$ consists of three weighted terms corresponding the image $\ImageErrors(\MaskedImages, \MaskedRenderedImages)$, depth $\DepthErrors(\MaskedDepths, \MaskedRenderedDepths)$, and $\MasksErrors(\EstimatedMasks, \RenderedMasks)$ mean absolute errors:
\begin{equation}\label{eq:sceneLoss}
    \SceneLoss =\sum_{\ObjectArg=1}^{\EstimatedNumObjects} (
        \SceneImageWeight \ImageErrors +
        \SceneDepthWeight \DepthErrors +
        \SceneMasksWeight \MasksErrors).
\end{equation}
\begin{figure}[th]
    \vspace{0.1cm}
    \centering
    \resizebox{0.945\columnwidth}{!}{%
    \input{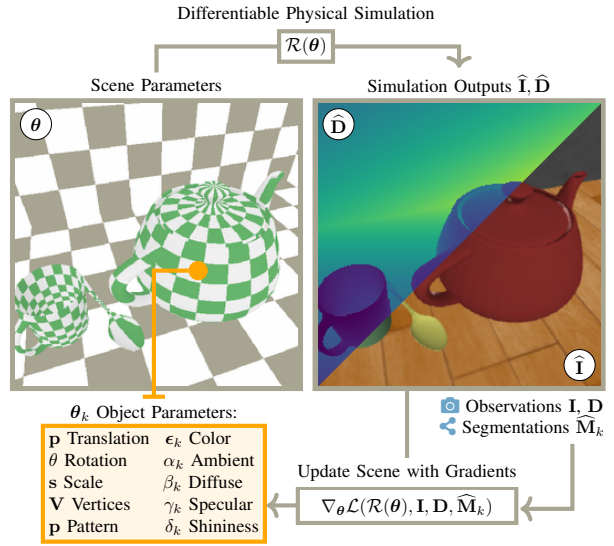}
    }
\caption{\protect\highlightOne{Differentiable rendering pipeline. The renderer $\DifferentiableRenderer$ takes scene parameters $\Parameters_{\ObjectArg}$ (left) to output RGBD images (right). The error between rendered and real observations is differentiated to provide gradients for optimizing $\Parameters_{\ObjectArg}$.}}~\label{fig:RenderPipeline}
    \vspace{-0.8cm}
\end{figure}
Let $\preceq$ denote the element-wise smaller than between two vectors and $\mathbf{u}_b=[\mathbf{1}, \Shininess_\text{max}]$ denote the upper bounds for the material parametrization $\Materials$ respectively.
Then, our inverse optimization problem reads:
\begin{align}
    \OptimizedSceneParameters =
        \argmin\limits_{\SceneParameters}
            \left\{
                \SceneLoss
            \right\}
            \text{ s.t.}
            \begin{cases}
                \Positions = t \Direction + \CameraOrigin \\
                \mathbf{0} \preceq \Materials \preceq \mathbf{u}_b
                \label{eq:sceneMathematicalOptimization}
            \end{cases}
\end{align}
\highlightOne{We use \LBFGS~to estimate all parameters $\OptimizedSceneParameters$.
    While \LBFGS~is known to be highly efficient is also prone to local minima, our methodology is explicitly designed to overcome this sensitivity.
    The ellipsoid initialization provides the necessary robustness against this sensitivity by ensuring the optimizer begins in the correct solution basin.
    The 3D line constraint then further stabilizes this convergence.
    This strategy is validated in Sec.~\mbox{\ref{sec:results_clvr_pose}}, where a baseline initialization fails to converge.
The optimization is run sequentially.
    We first optimize the light and floor, then the sphere locations, shapes, and materials.
    This approach converges in under 300 steps, reducing optimization steps by 2.3x compared to other methods applied to analogous scenes~{\cite{arriaga2024bayesian}}.
    A visualization of the optimization is provided in Figure~{\ref{fig:OptimizationStepsCLVRPOSE}}.
}

        \subsection{Differentiable Mesh Rendering}
\begin{figure*}[ht]
    \input{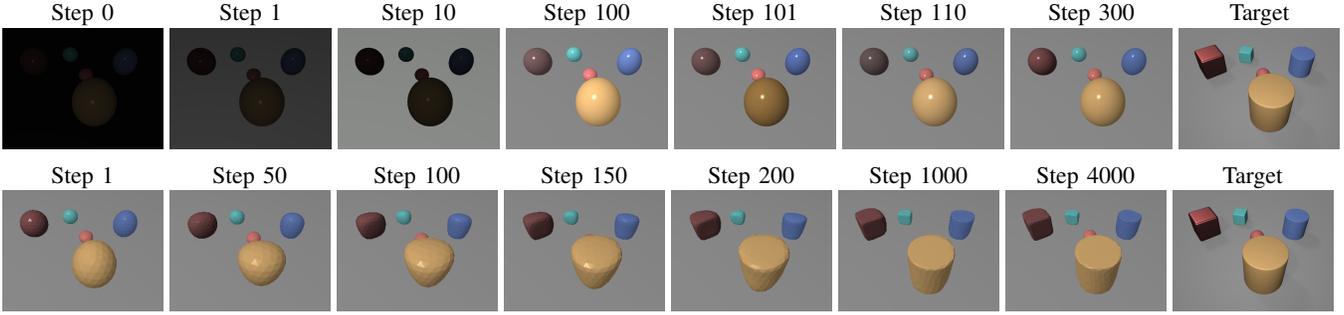}
    \caption{Results of the inverse optimization steps using the differentiable renderer, achieving accurate scene reconstruction from a single \RGBD~image. Top row) Scene optimization from Eq.~\ref{eq:sceneMathematicalOptimization}. Bottom row) Mesh optimization from Eq.~\ref{eq:meshMathematicalOptimization}.~\label{fig:OptimizationStepsCLVRPOSE}}
    \vspace{-0.5cm}
\end{figure*}

            The final step optimizes meshes to fit more complex geometries.
We render each object shape using a mesh $\Mesh$ whose deformation is driven by a control cage $\ShapeParameters \in \Reals^{\NumCageVertices \times 3}$ using mean value coordinates~\cite{ju2023mean}:
\begin{align}
    \RenderedImage, \RenderedDepth, \RenderedMasks = \DifferentiableRenderer(\ShapeParameters).
\end{align}
In this optimization step, we include three regularization terms, namely a smooth mesh loss $\SmoothMeshLosses$, consisting of the discrete Laplacian regularizer~\cite{nicolet2021large}, a disparity smoothness depth loss $\SmoothDepthLosses$~\cite{godard2017unsupervised}, and a newly introduced volume loss $\VolumeLosses=\vert \SceneVolumes - \MeshVolumes \vert^{2}$, enforcing that the predicted spherical volumes  $\SceneVolumes$ from the previous optimization step remain close to the volume of the predicted meshes $\MeshVolumes$.
Thus, the mesh loss $\MeshLoss$ consists of a scene loss plus the regularization terms:
\begin{align}\label{eq:MeshLoss}
    \MeshLoss =
        \SceneLoss +
        \sum_{\ObjectArg=1}^{\NumObjects}
        (\SmoothDepthLossWeight \SmoothDepthLosses +
         \SmoothMeshLossWeight  \SmoothMeshLosses +
         \VolumeLossWeight      \VolumeLosses).
\end{align}
Consequently, we optimize the meshes by
\begin{align}
    \OptimizedShapeParameters = \argmin_{\ShapeParameters} \MeshLoss(\Image, \Depth, \EstimatedMasks, \DifferentiableRenderer(\ShapeParameters)).~\label{eq:meshMathematicalOptimization}
\end{align}
The optimization problem is solved using ADAMW~\cite{loshchilov2017fixing} over 4K steps using a learning rate of $5e^{-3}$.
The bottom row of Figure~\ref{fig:OptimizationStepsCLVRPOSE} shows the optimization results.
The object's pose is assigned by applying PCA to the reconstructed mesh.

        \section{Differentiable Renderer}
            \highlightOne{We developed a differentiable ray tracer for triangular meshes using only \JAX~primitives, which enables GPU-based optimization with respect to the scene parameters shown in Figure~{\ref{fig:RenderPipeline}}.
All benchmarks were run on an RTX 5090 GPU.
Using the Phong reflection model~{\cite{bui1975}}, the 15K-vertex scene in Figure~{\ref{fig:RenderPipeline}} renders a $512 \times 512$ image in \mbox{\qty{0.97}{\s}}.
To validate our renderer, including its specialized analytic sphere (JAX-sphere) and general mesh (JAX-mesh) components, we benchmarked it against the PyTorch3D rasterizer and the Mitsuba path tracer by rendering a $512 \times 512$ image of a sphere with 162 vertices, matching our object experiments' vertex count.
The JAX-sphere renderer is exceptionally efficient (0.12ms), 32x faster than PyTorch3D (3.9ms) and 239x faster than Mitsuba (29.1ms).
JAX-mesh (7.8ms) is 3.7x faster than Mitsuba and 2x slower than PyTorch3D, balancing ray-tracing fidelity with performance.}

            Furthermore, as previously shown in Eq.~\ref{eq:sceneLoss} and Eq.~\ref{eq:MeshLoss}, our model optimization uses the predicted masks $\EstimatedMasks$. 
However, rendering binary masks $\RenderedMasks$ is inherently a non-differentiable procedure, as it assigns binary values to each pixel.
Previous work developed soft-mask functions for triangular meshes; however, these methods require recomputing the K-nearest neighbors between each pixel and the projected mesh faces~\cite{liu2019soft,ravi2020pytorch3d}.
Our work simplifies this methodology by introducing a soft-mask function operating directly on rendered depth images; thus, eliminating the need for pixel-mesh correspondence.
The soft-mask function starts by computing a min-max normalization centered at 0, using the minimum $\MinDepth$ and maximum $\MaxDepth$ observable depths.
Moreover, rendering functions typically assign zero depth values to distant points.
These points are then assigned a constant negative depth value $c$.
Finally, we apply a sigmoid function scaled by a constant $\kappa$ to control the mask activation threshold.
Thus for each depth image pixel $d_{u,v}$ we compute the soft-mask values $m_{u,v}$:
\begin{equation}\label{eq:minmaxnorm}
    d^{'}_{u,v} = \left\{
        \begin{array}{cl}
            \dfrac{- d_{u,v} + \MaxDepth}{\MaxDepth - \MinDepth} - 0.5 & \textrm{if} \; d_{u,v} > \epsilon \\
            c & \textrm{otherwise}
        \end{array}
    \right.
\end{equation}
\begin{equation}
    m_{u,v} = \textrm{sigmoid}(\kappa \cdot d^{'}_{u,v})~\label{eq:sigmoidScaled}
\end{equation}
All constant values and the full implementation of our differentiable renderer will be made available open-source.

    \section{Results}\label{sec:results}
        To maximize generalization across diverse scene distributions, we built a benchmark dataset using scenes and objects from CLVR~\cite{johnson2017clevr}.
After demonstrating strong performance on this benchmark, we assessed the generalization capability of our model in a zero-shot setting on three standard datasets: $\FEWSOL$~\cite{chao2023fewsol}, $\MOPED$~\cite{park2020latentfusion}, and $\LINEMOD$~\cite{brachmann2014learning}, without using any training data or modifying the model or its hyperparameters.
Finally, we deployed our model within a robotic system and evaluated its zero-shot grasping performance using objects from the $\YCB$ dataset~\cite{calli_ycb}.
\begin{figure}[H]
    \vspace{0.30cm}
    \centering
    \begin{subfigure}{0.9\columnwidth}
        \input{sections/images/image_pose_results.tex}
    \end{subfigure}
\caption{\protect\highlightOne{Zero-shot pose estimation results in $\FEWSOL$, $\CLVRPOSE$, $\MOPED$ and $\LINEMOD$. All objects are previously unseen by the model.}~\label{fig:PoseEstimationResults}}
    \vspace{-0.5cm}
\end{figure}
We evaluate pose accuracy using the Visible Surface Discrepancy Average Recall ($AR_{VSD}$), a standard metric within the BOP benchmark for assessing visible surface alignment~\cite{hodavn2020bop}.
Unlike mesh-dependent metrics, $AR_{VSD}$ can be applied without ground truth meshes, enabling its use on datasets like $\FEWSOL$.
However, the $AR_{VSD}$ reliance on visibility masks makes it sensitive to segmentation errors~\cite{hodan2016evaluation}.
To ensure robust evaluation, we supplement $AR_{VSD}$ with the geometry-based Chamfer and Hausdorff distances.
These directly compare point clouds and thus provide segmentation-independent pose metrics less susceptible to mask inaccuracies.
\begin{figure}[ht]
    \centering
    \includegraphics[width=0.90\columnwidth]{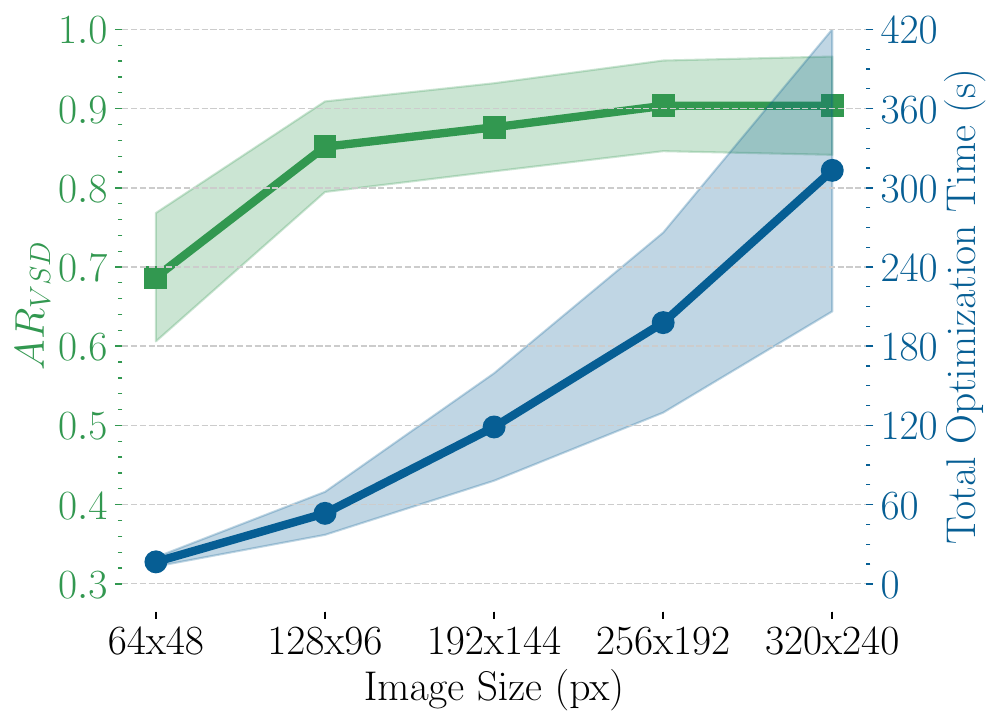}
    \caption{\protect\highlightOne{Tradeoff between performance ({$AR_{VSD}$}), total optimization time and image resolution. Higher resolutions yield better results at the cost of longer a optimization.}}~\label{fig:meshes_resolution_tradeoff}
    \vspace{-0.95cm}
\end{figure}

        \subsection{CLVR Benchmark}\label{sec:results_clvr_pose}
\begin{figure}[ht]
    \input{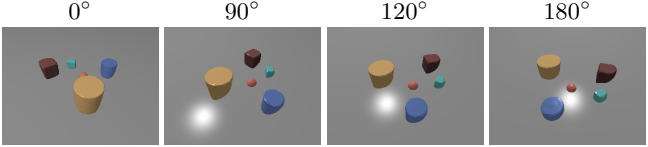}
    \caption{Reconstructed scene at different angles. Despite the backside being unobserved, we achieve a consistent reconstruction by leveraging our physical model and priors.\label{fig:BackCLVRPOSE}}
    \vspace{-0.3cm}
\end{figure}

\begin{figure}[ht]
    \centering
    \includegraphics[width=0.85\columnwidth]{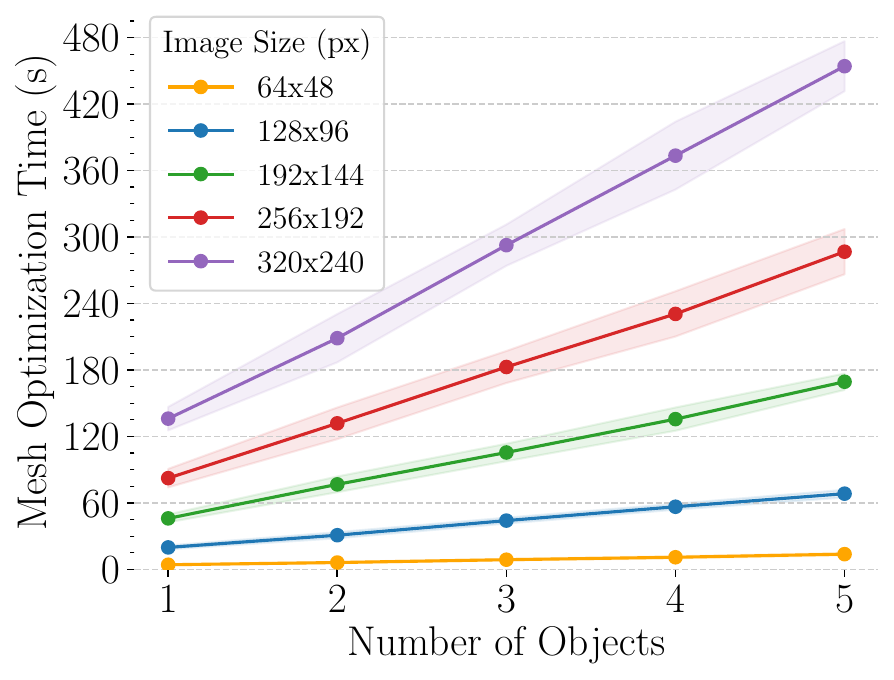}
    \caption{\protect\highlightOne{Mesh optimization times with varying number of objects at different resolutions. At every resolution the optimization time increase linearly with the number of objects.}~\label{fig:meshes_time_scaling}}
    \vspace{-0.6cm}
\end{figure}

            \highlightOne{We introduce $\CLVRPOSE$, a CLEVR-based dataset with 50 scenes, each featuring one to five unique object instances with 6D pose labels.
    These instances combine various colors, shapes (sphere, cube, cylinder), sizes (3cm, 5cm, 7cm), and materials (rubber, metal).
    Despite its visually simple appearance, \CLVRPOSE~is designed to test generalization under limited data, presenting open challenges such as textureless surfaces, symmetries, and reflective materials.
    We analyzed model performance using 500 runs on an RTX 5090 GPU.
    These runs validated that ellipsoid initialization (Eq.~{\ref{eq:ellipsoidMathematicalOptimization}}) is critical, achieving $0.656 \pm 0.05$ $AR_{VSD}$ on its own, while a baseline (using Normal and Truncated Normal for $\Position$ and $\Scale$ fit to dataset statistics) failed to converge ($0.008 \pm 0.04$ $AR_{VSD}$).
    Moreover, the 3D line constraint offered a minor stabilization ($0.653 \pm 0.06$ vs. $0.648 \pm 0.05$ $AR_{VSD}$).
    Figure~{\ref{fig:meshes_resolution_tradeoff}} shows how higher image resolutions improve $AR_{VSD}$~at the cost of total optimization time.
    This total time sums the ellipsoid (Eq.~{\ref{eq:ellipsoidMathematicalOptimization}}), scene (Eq.~{\ref{eq:sceneMathematicalOptimization}}), and mesh (Eq.~{\ref{eq:meshMathematicalOptimization}}) optimizations. The ellipsoid and scene stages take respectively $2.76 \pm 1.01$s and $4.73 \pm 0.31$s, while the mesh optimization time varies with resolution and object count (Figure~{\ref{fig:meshes_resolution_tradeoff}} and Figure~{\ref{fig:meshes_time_scaling}}).
    Figure~{\ref{fig:BackCLVRPOSE}} visualizes the scene reconstruction from multiple unseen views.
}
\vspace{-0.65cm}

        \subsection{FEWSOL Results}
            \begin{figure}[!t]
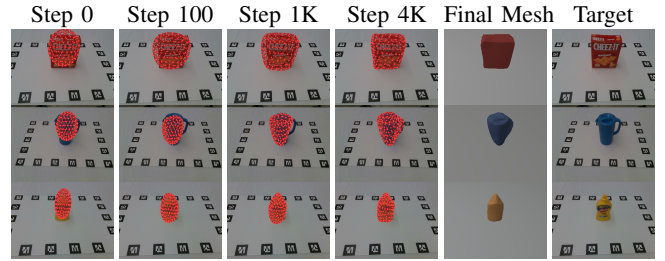

    \begin{subfigure}{\columnwidth}
        \input{sections/images/fewsol/mesh_optimization_00_00.tex}

        \input{sections/images/fewsol/mesh_optimization_01_01.tex}

        \input{sections/images/fewsol/mesh_optimization_03_00.tex}
    \end{subfigure}
    \caption{Zero-shot reconstructions for three \FEWSOL~objects at different mesh optimization steps of Eq.~\ref{eq:meshMathematicalOptimization}.\label{fig:invFEWSOL}}
    \vspace{-0.7cm}
\end{figure}
We evaluated our model's performance on 20 objects in the $\FEWSOL$ dataset using $AR_{VSD}$, Chamfer distance, and Hausdorff distance.
\begin{figure*}[htpb]
    \centering
    \includegraphics[width=0.975\textwidth]{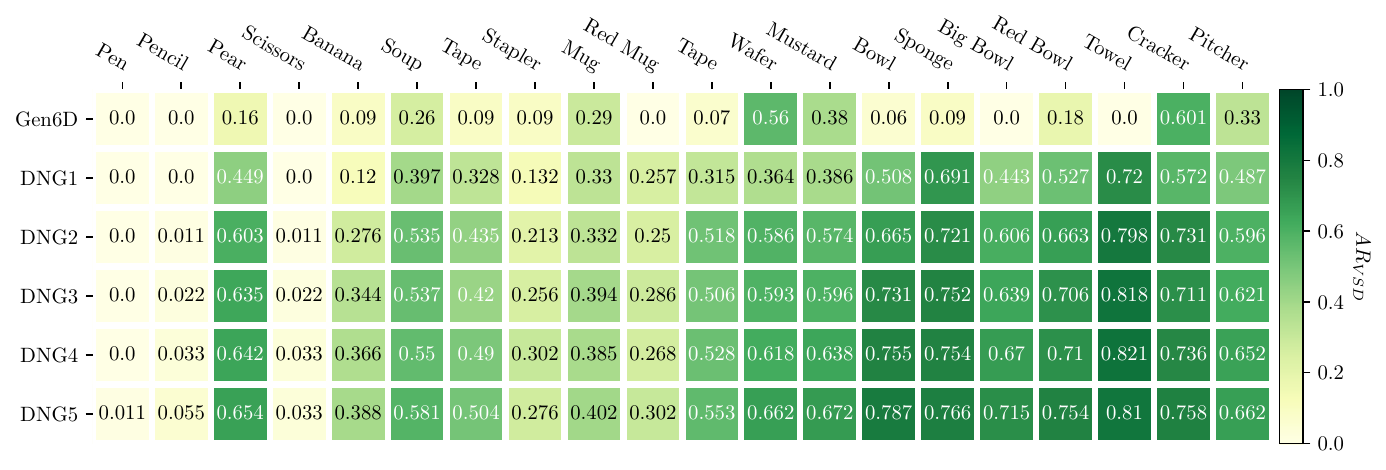}
    \captionsetup{skip=-7pt} 
    \caption{$AR_{VSD}$ heatmap for \FEWSOL~objects ordered left-to-right by increasing size, comparing \GenSixD~with \MyModel~variants at increasing resolution (top-to-bottom). \MyModel~yields higher performance, improving with object size and input resolution.~\label{fig:HeatmapFEWSOLVSD}}
    \vspace{-0.25cm}
\end{figure*}
\begin{table*}[ht]
    \normalsize
    \captionsetup{skip=3pt} 
    \caption{Chamfer and Hausdorff distances for \FEWSOL~objects, comparing \GenSixD, \OnePosePlusPlus, and our variant \MyModelTwo. Our low-resolution model outperforms the baselines in terms of reconstruction scores ($\downarrow$ lower is better).~\label{table:ResultsFEWSOL}}
    \resizebox{\textwidth}{!}{
        \begin{tabular}{c c c c c c c c c c c c c c c c c c c c c}
            & \rot{\textbf{Pen}} & \rot{\textbf{Pencil}} & \rot{\textbf{Pear}} & \rot{\textbf{Scissors}} & \rot{\textbf{Banana}} & \rot{\textbf{Soup}} & \rot{\textbf{Tape}} & \rot{\textbf{Stapler}} & \rot{\textbf{Mug}} & \rot{\textbf{Red Mug}} & \rot{\textbf{Tape}} & \rot{\textbf{Wafer}} & \rot{\textbf{Mustard}} & \rot{\textbf{Bowl}} & \rot{\textbf{Sponge}} & \rot{\textbf{Big Bowl}} & \rot{\textbf{Red Bowl}} & \rot{\textbf{Towel}} & \rot{\textbf{Cracker}} & \rot{\textbf{Pitcher}} \\
            \cline{2-21} \noalign{\vskip\doublerulesep \vskip-\arrayrulewidth} \cline{2-21}
            \multicolumn{1}{c}{\textbf{Model}} & \multicolumn{20}{c}{\textbf{Chamfer Distance} ($10^3$) $\downarrow$} \\
            \cline{2-21} \noalign{\vskip\doublerulesep \vskip-\arrayrulewidth} \cline{2-21}
            \OnePosePlusPlus          &  68.31 & 18.78 & 13.76 & 33.39 & 35.47 & 9.89 & 15.25 & 4.86 & 9.34 & 16.66 & 7.74 & 25.45 & 23.73 & 7.07 & 2.17 & 4.77 & 5.62 & 4.99 & 19.72 & 13.45 \\
            \GenSixD              & 4.56 & 1.89 & 20.51 & 2.22 & 4.99 & 1.81 & 106.89 & 16.63 & 3.41 & 4.51 & 65.95 & 26.82 & 3.84 & 34.44 & 2.77 & 62.34 & 34.95 & 44.84 & 2.52 & 8.31 \\
            \MyModelTwo~(ours) & \textbf{0.37} & \textbf{0.66} & \textbf{0.35} & \textbf{0.45} & \textbf{0.22} & \textbf{0.66} & \textbf{0.71} & \textbf{0.90} & \textbf{0.74} & \textbf{0.76} & \textbf{0.27} & \textbf{1.01} & \textbf{1.27} & \textbf{0.33} & \textbf{0.46} & \textbf{0.68} & \textbf{0.37} & \textbf{0.22} & \textbf{1.62} & \textbf{2.33} \\
            \multicolumn{1}{c}{} & \multicolumn{20}{c}{\textbf{Hausdorff Distance $\downarrow$}} \\
            \cline{2-21} \noalign{\vskip\doublerulesep \vskip-\arrayrulewidth} \cline{2-21}
            \OnePosePlusPlus          & 0.313 & 0.225 & 0.157 & 0.248 & 0.241 & 0.158 & 0.18 & 0.156 & 0.146 & 0.188 & 0.151 & 0.285 & 0.299 & 0.158 & 0.067 & 0.124 & 0.179 & 0.114 & \textbf{0.195} & 0.154 \\
            \GenSixD              & 0.239 & 0.089 & 0.139 & 0.147 & 0.118 & 0.107 & 0.319 & 0.154 & 0.296 & 0.111 & 0.269 & 0.204 & 0.192 & 0.195 & 0.132 & 0.249 & 0.255 & 0.237 & 0.238 & 0.364 \\
            \MyModelTwo~(ours) & \textbf{0.040} & \textbf{0.047} & \textbf{0.057} & \textbf{0.066} & \textbf{0.039} & \textbf{0.084} & \textbf{0.052} & \textbf{0.064} & \textbf{0.075} & \textbf{0.067} & \textbf{0.053} & \textbf{0.110} & \textbf{0.127} & \textbf{0.070} & \textbf{0.073} & \textbf{0.110} & \textbf{0.097} & \textbf{0.048} & 0.232 & \textbf{0.239} \\
        \end{tabular}
        }
        \vspace{-0.5cm}
    \end{table*}
To analyze the impact of resolution in real images, we compare variants of \MyModel~that differ only in input image size.
Each variant processes images at a fixed fraction of the original resolution ($640 \times 480$); for example, \MyModelTwo~uses 20\% (i.e., $128 \times 96$), while \MyModelThree~uses 30\%, and so on.
We evaluate these variants against state-of-the-art, model-free methods, \GenSixD~and \OnePosePlusPlus.
As shown in Figure~\ref{fig:HeatmapFEWSOLVSD} and Table~\ref{table:ResultsFEWSOL}, \MyModel~variants consistently outperform both baselines, with accuracy increasing alongside object size.
The $AR_{VSD}$ metric was omitted for \OnePosePlusPlus~because substantial inaccuracies in the mask predictions resulted in correspondingly low scores.

        \subsection{MOPED Results}
            We further evaluated our models on the \MOPED~dataset.
The performance of \GenSixD~and \OnePosePlusPlus~is summarized in Table~\ref{table:ResultsMOPED}, where we compare $AR_{VSD}$, Chamfer Distance, and Hausdorff Distance.
Our method consistently outperforms both \GenSixD~and \OnePosePlusPlus~across most evaluation metrics.
Additional visualizations of the \MOPED~predictions are displayed in Figure~\ref{fig:PoseEstimationResults}.
\begin{table}[ht]
    \centering
    \normalsize
    \captionsetup{skip=3pt} 
    \caption{Evaluation on the \MOPED~dataset with Chamfer and Hausdorff distance, and $AR_{VSD}$ ($\uparrow$ higher is better).\label{table:ResultsMOPED}}
    \resizebox{0.85\columnwidth}{!}{%
    \begin{tabular}{c c c c c c c}
        & \textbf{Cheezit} & \textbf{Plane} & \textbf{Pouch} & \textbf{Duplo} & \textbf{Rinser} & \textbf{Mug} \\[0.8ex]
        \cline{2-7} \noalign{\vskip\doublerulesep \vskip-\arrayrulewidth} \cline{2-7}
        \multicolumn{1}{c}{\textbf{Model}} & \multicolumn{6}{c}{\textbf{$AR_{VSD}$ $\uparrow$}} \\
        \cline{2-7} \noalign{\vskip\doublerulesep \vskip-\arrayrulewidth} \cline{2-7}
        Gen6D              & 0.145 & 0.004 & 0.110 & 0.056 & 0.143 & 0.073 \\
        \MyModelTwo~(ours) & \textbf{0.656} & \textbf{0.286} & \textbf{0.735} & \textbf{0.450} & \textbf{0.533} & \textbf{0.136} \\
        \multicolumn{1}{c}{} & \multicolumn{6}{c}{\textbf{Chamfer Distance} ($10^3$) $\downarrow$} \\
        \cline{2-7} \noalign{\vskip\doublerulesep \vskip-\arrayrulewidth} \cline{2-7}
        OnePose++          & 22.5 & 06.2 & 11.9 & 11.2 & 49.1 & 10.9 \\
        Gen6D              & 8.7  & 23.8 & 10.3 & 15.9 & 11.7 & 10.3  \\
        \MyModelTwo~(ours) & \textbf{1.38} & \textbf{2.64} & \textbf{0.49} & \textbf{1.41} & \textbf{0.52} & \textbf{2.72} \\
        \multicolumn{1}{c}{} & \multicolumn{6}{c}{\textbf{Hausdorff Distance $\downarrow$}} \\
        \cline{2-7} \noalign{\vskip\doublerulesep \vskip-\arrayrulewidth} \cline{2-7}
        OnePose++          & 0.242 & \textbf{0.127} & 0.166 & 0.176 & 0.297 & 0.135 \\
        Gen6D              & 0.234 & 0.364 & 0.171  & 0.225 & 0.147 & \textbf{0.131} \\
        \MyModelTwo~(ours) & \textbf{0.136} & 0.163 & \textbf{0.073} & \textbf{0.137} & \textbf{0.065} & 0.132
    \end{tabular}
    }
    \vspace{-0.75cm}
\end{table}

        \subsection{LINEMOD-OCCLUDED Results}
            To assess the robustness of our model in cluttered and occluded environments, we evaluate it on the $\LINEMOD$ dataset, using a subset with all objects present to focus on multi-object real-world scenes.
Our method achieves a mean $AR_{VSD}$ score of $0.275 \pm 0.083$, with qualitative results presented in Figure~\ref{fig:PoseEstimationResults}.
Notably, it ranks among the top 100 entries on the $\LINEMOD$ benchmark of the \href{https://bop.felk.cvut.cz/leaderboards/pose-estimation-bop19/lm-o/}{BOP Challenge leaderboard}~\cite{hodavn2020bop}, despite competing against fully supervised models trained on large real and synthetic datasets with access to 3D object models.
In contrast, our approach is entirely zero-shot, requiring no training data, object models, or dataset-specific tuning.
Beyond pose estimation, we also evaluate the predicted segmentation under occlusions.
SAM achieved a mean IoU of 0.712 ($\pm$0.072) between predicted and ground-truth masks.
These results demonstrate that SAM can estimate accurate predictions even under large occlusions.

        \subsection{Zero-shot Robot Grasping}
\begin{figure}[ht]
    \centering
    \scalebox{0.88}{
    \begin{tikzpicture}
        \tikzset{
          /pgf/arrow keys/Centered Circle size/.style={length={#1}, width'=+0pt +1, sep=-0.5*(#1)},
          Centered Circle/.tip={Circle[Centered Circle size=6pt]},
          robotSetupLine/.style={thick, Centered Circle-., robotSetupLineColor, very thick}
        }

        \node[anchor=south west,inner sep=0] (image) at (0,0) {\includegraphics[width=10cm]{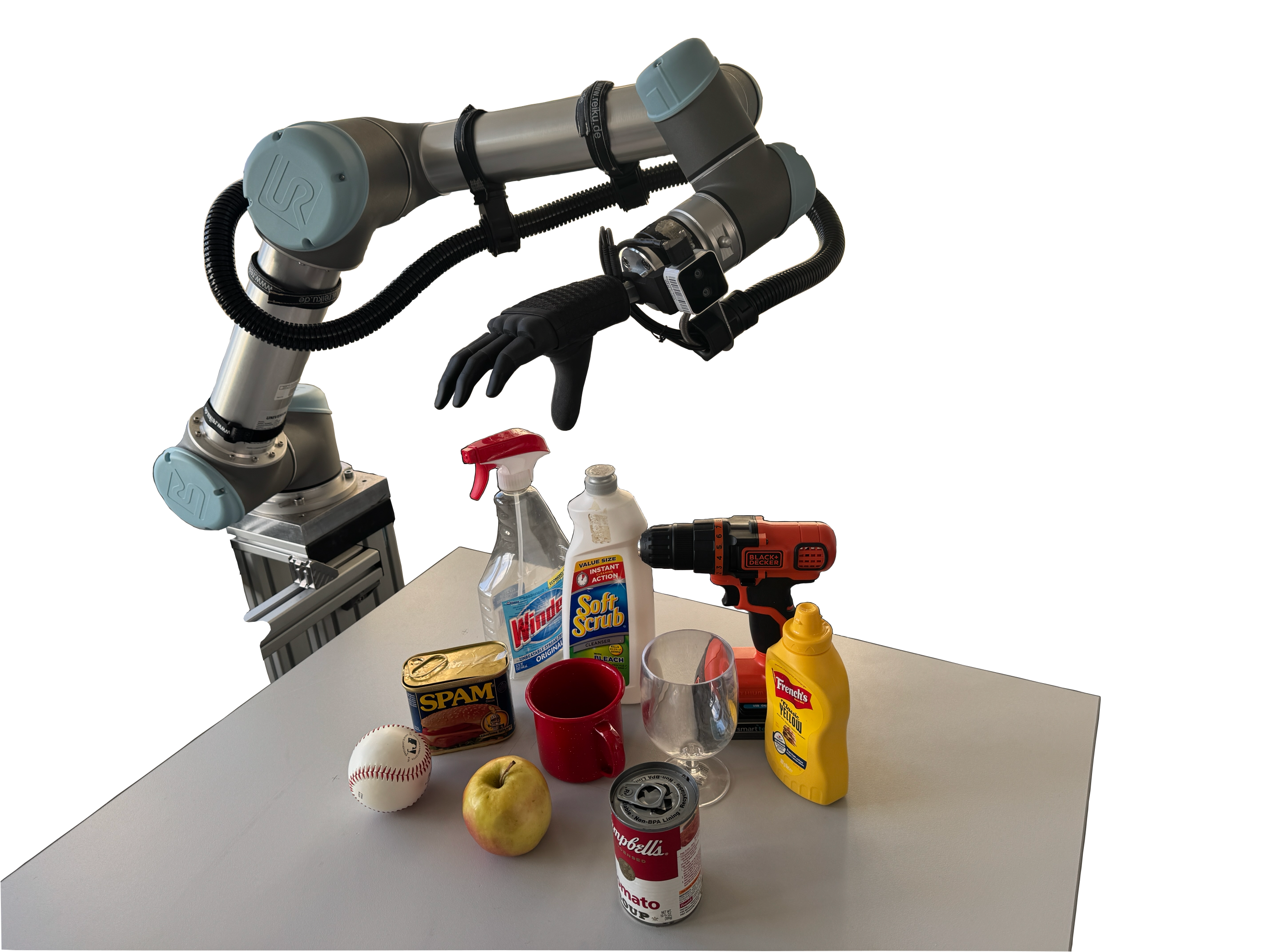}};

        \begin{scope}[x={(image.south east)},y={(image.north west)}]
            \draw[robotSetupLine, line width=2pt] (0.58, 0.72) -- (0.67, 0.80) node[right, align=center, text=robotSetupTextColor] {D405 RGBD \\ Camera};
            \draw[robotSetupLine, line width=2pt] (0.49, 0.35) -- (0.67, 0.53) node[right, align=center, text=robotSetupTextColor] {YCB Objects};
            \draw[robotSetupLine, line width=2pt] (0.45, 0.70) -- (0.33, 0.92) node[above, align=center, text=robotSetupTextColor] {Mia Hand (4 DoF)};
            \draw[robotSetupLine, line width=2pt] (0.20, 0.70) -- (0.08, 0.75) node[above, align=center, text=robotSetupTextColor] {UR5 \\ (6 DoF)};
        \end{scope}
    \end{tikzpicture}
}
    \caption{Zero-shot grasping setup consisting of a UR5 robotic arm equipped with a Mia hand and a RealSense D405 camera. The table displays our 10 tested \YCB~dataset objects.~\label{fig:robot_setup}}
\end{figure}
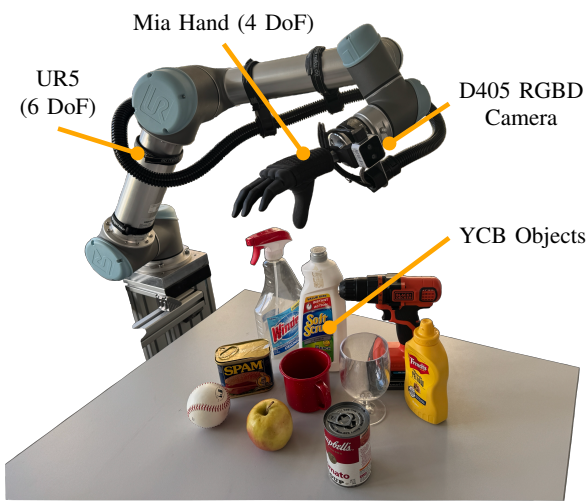

            To validate our zero-shot reconstruction algorithm, we conducted 224 zero-shot grasping tests on 10 diverse $\YCB$ objects, achieving a total success rate of 89.28\%.
As shown in Figure {\ref{fig:robot_setup}}, our experimental setup includes a 6-DoF UR5 arm, a 4-DoF Mia Hand, and a RealSense D405 $\RGBD$ camera.
The pipeline consists of capturing an $\RGBD$ image from a fixed viewpoint, followed by our 3D reconstruction methodology, multiple simulated grasps are evaluated, and a successful grasp is selected and executed on the real robot.
To ensure diverse evaluation, we selected 10 $\YCB$ objects varying in size, weight, and transparency, shown in Figure~\ref{fig:robot_setup}.
For axisymmetric objects (baseball, apple, tomato soup can, wine glass), we performed 20 grasps trials at different table locations.
For non-axisymmetric objects, we conducted tests at 6 positions with 4 in-plane rotations ($\ang{0}, \ang{45}, \ang{90}, \ang{135}$), covering a $\qtyproduct{600 x 600}{\milli\metre}$ area in the robot’s workspace.
For each trial, grasp orientations were sampled around the object’s upright z-axis in the range $[\ang{10}, \ang{120}]$, selecting the first successful attempt.
\highlightOne{The front-to-back sampling order prioritizes the more accurate object fronts, as these have lower uncertainty~{\cite{arriaga2024bayesian}}.
While more complex policies exist, our straightforward approach validates our zero-shot reconstruction for grasping novel objects without additional test-time data, achieving an $89.3\%$ success rate (Table~{\ref{tab:success_rate}}).
    For comparison, GraspSAM~{\cite{noh2024graspsam}}, which also leverages SAM, achieves a comparable 86\% success rate on 20 similar household objects in cluttered scenes. However, their data-driven approach is pre-trained on \textasciitilde540 million labeled grasps, whereas our reconstruction-based method requires no grasp-specific training data.}
\vspace{-0.25cm}
\begin{table}[h]
    \centering
    \captionsetup{skip=3pt} 
    \caption{Zero-shot grasping accuracy using our scene reconstruction algorithm across a diverse set of $\YCB$ objects.~\label{tab:success_rate}}
    \resizebox{\columnwidth}{!}{%
    \begin{tabular}{l c c c c c c c c c c | c}
        & \rotb{\textbf{ball}} & \rotb{\textbf{apple}} & \rotb{\textbf{soup}} & \rotb{\textbf{glass}} & \rotb{\textbf{mustard}} & \rotb{\textbf{red cup}} & \rotb{\textbf{bleach}} & \rotb{\textbf{spam}} & \rotb{\textbf{cleaner}} & \rotb{\textbf{drill}} & \rotb{\textbf{Total}} \\
        \cline{2-12} \noalign{\vskip\doublerulesep \vskip-\arrayrulewidth} \cline{2-12}
        Attempts & 20 & 20 & 20 & 20 & 24 & 24 & 24 & 24 & 24 & 24 & \textbf{224} \\
        Success & 20 & 20 & 20 & 19 & 23 & 23 & 20 & 20 & 19 & 16 & \textbf{200} \\
        Accuracy (\%) & 100 & 100 & 100 & 95 & 95.8 & 95.8 & 83.3 & 83.3 & 79.2 & 75 & \textbf{89.3} \\
    \end{tabular}%
    }
    \vspace{-0.40cm}
\end{table}

    \section{Conclusion}\label{sec:conclusion}
        To address the robotic perception challenges of sample complexity, interpretability, and extensive test-time data, we introduce a \modelname~model that combines a foundation segmentation model with a physics-based differentiable renderer for performing zero-shot scene reconstruction without any additional 3D training data.
\highlightOne{This capability enables a robotic system to grasp novel objects without any prior 3D models or annotated test-time samples.
    Current limitations include a total optimization time of approximately 1 minute with the baseline model \MyModelTwo~on an RTX 5090 GPU, and the reliance of bounding boxes.
    Future work aims to accelerate inference through learned priors, and integrate foundation models for object detection.
}

    \bibliographystyle{IEEEtran}
    \bibliography{references}

\end{document}